\documentclass{styles/svproc}
\usepackage{lipsum}
\usepackage[most]{tcolorbox}
\usepackage{float}
\usepackage{hyperref}
\usepackage[capitalise]{cleveref}
\usepackage{todonotes}

\usepackage{url}

\begin{document}
\mainmatter              %
\title{VoyagerVision: Investigating the Role of Multi-modal Information for Open-ended Learning Systems}
\titlerunning{Voyager Vision}  %
\author{Ethan Smyth \and Alessandro Suglia}
\authorrunning{Ethan Smyth \and Alessandro Suglia} %
\tocauthor{}
\institute{School of Mathematical and Computer Sciences,\\ Heriot-Watt University, Edinburgh, UK,\\
\email{a.suglia@hw.ac.uk},\\ 
\texttt{http://www.macs.hw.ac.uk/InteractionLab}
}
\maketitle              %

\begin{abstract}
Open-endedness is an active field of research in the pursuit of capable Artificial General Intelligence (AGI), allowing models to pursue tasks of their own choosing. Simultaneously, recent advancements in Large Language Models (LLMs) such as GPT-4o \cite{gpt4o} have allowed such models to be capable of interpreting image inputs. Implementations such as OMNI-EPIC \cite{omni_epic} have made use of such features, providing an LLM with pixel data of an agent's POV to parse the environment and allow it to solve tasks. This paper proposes that providing these visual inputs to a model gives it greater ability to interpret spatial environments, and as such, can increase the number of tasks it can successfully perform, extending its open-ended potential. To this aim, this paper proposes VoyagerVision - a multi-modal model capable of creating structures within Minecraft using screenshots as a form of visual feedback, building on the foundation of Voyager. VoyagerVision was capable of creating an average of 2.75 unique structures within fifty iterations of the system, as Voyager was incapable of this, it is an extension in an entirely new direction. Additionally, in a set of building unit tests VoyagerVision was successful in half of all attempts in flat worlds, with most failures arising in more complex structures. Project website is available at \url{https://esmyth-dev.github.io/VoyagerVision.github.io/}.

\keywords{Multimodal Generative AI, Embodied AI, Vision-and-Language Models, Open-ended Learning Systems}
\end{abstract}

\section{Introduction}
\label{sec:intro}

Artificial general intelligence (AGI)—proficiency across a wide range of cognitive tasks including metacognition \cite{RN10}—has the potential for profound societal impact. For instance, it’s estimated that LLMs could affect at least 10\% of work tasks for 80\% of the U.S. workforce \cite{labor}, illustrating the influence of today’s "Emergent AGI" \cite{RN10}. More advanced AGI, particularly with superhuman performance, would likely bring even greater disruption. Open-endedness is increasingly viewed as essential to achieving such Artificial Superhuman Intelligence (ASI) \cite{essential}. We consider testing agents in this open-ended setup as a way to truly assess their capability to generalise to unseen object instances~\cite{suglia2020compguesswhat}, unseen tasks or open worlds~\cite{fan2022minedojo}---an essential skill for robust AI models that are able to operate in dynamic real-world environments. 

\begin{figure}
    \centering
    \includegraphics[width=1\linewidth]{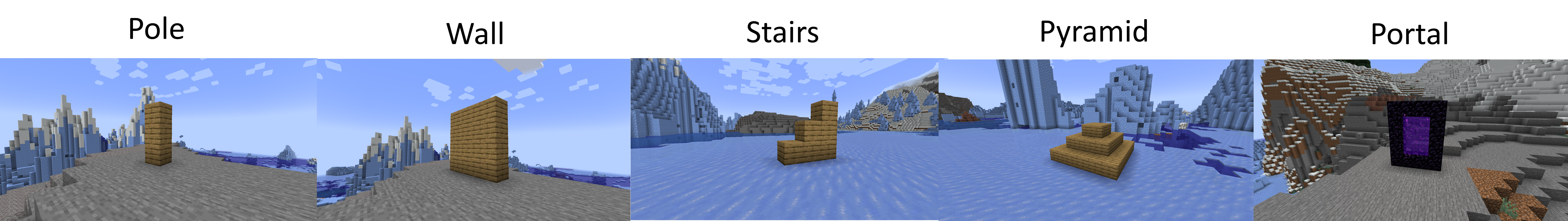}
    \caption{In this paper, we present VoyagerVision, the first Embodied Multimodal AI agent able to learn to solve both resource gathering tasks as well as novel building tasks such as poles, walls, and pyramids.}
    
    \label{fig:allBuilds}
\end{figure}

This work explores Voyager \cite{voyager}, an open-ended Minecraft agent initially limited to text inputs due to prior LLM constraints. As such, motivating this project to enhance Voyager with screenshot inputs to integrate the pursuit of building-focused tasks.
Therefore, we propose VoyagerVision, an extension of the original Voyager framework that incorporates agent-perspective screenshots alongside textual inputs, thereby transforming it into a multimodal Minecraft agent. This enables the model to complete a new class of tasks, such as building structures (see \cref{fig:allBuilds}), that were previously infeasible due to the text-only nature of the system. To support image inputs, we replace the original language model with GPT-4o \cite{gpt4o} and implement several important changes to the prompts to support adaptations. %
A three-stage evaluation was conducted: first, GPT-4o was used without screenshots, revealing reduced efficiency and prompting challenges; second, screenshots were added, showing no additional performance degradation; and third, the full VoyagerVision system demonstrated the agent’s ability to build multiple novel structures, highlighting the value of multimodal integration. One of the experiments completed involved a set of unit tests where the agent was tasked with proving its ability to create a pre-defined structure. Each of the structures are shown above in \cref{fig:allBuilds}. The structures become more difficult to construct, moving from left to right. With the most difficult, the portal, involving both the construction and activation of a "nether portal" an in game creation which takes multiples steps to create and activate. 

\section{Related Work}
Open-endedness refers to the ability of a system to continually generate outputs that are both novel and learnable to an observer \cite{RN10}. Despite significant interest, truly open-ended systems remain rare. AI systems today can reach a superhuman level of performance, but only in a narrow scope, like AlphaZero \cite{AlphaZero}, or broader but less performant, as seen in modern large language models (LLMs)~\cite{grattafiori2024llama,gpt4o}.

Voyager \cite{voyager} presents an open-ended agent that explores Minecraft autonomously through code generation and self-critique. Its ability to continually learn and adapt in an open world environment makes it a promising example in this area, although it is still constrained by its lack of multi-modal capabilities, relying exclusively on textual inputs.

This limitation points to a broader gap in current approaches. Real-world open-ended agents must operate across modalities and scale to vast action spaces. Systems like OMNI and OMNI-EPIC \cite{omni,omni_epic} explore this by combining large vision and language models to tackle complex, multi-step tasks in rich environments. Building on this line of work, this project investigates how Voyager can be extended with visual understanding, bringing it closer to the kinds of agents that could eventually generalize to real-world tasks.

\subsection{Voyager}
\label{sec:voyager}
In this section, we present a detailed overview of the Voyager architecture that will provide important background to understand our extension Voyager Vision. 

\begin{figure}
    \centering
    \includegraphics[width=1\linewidth]{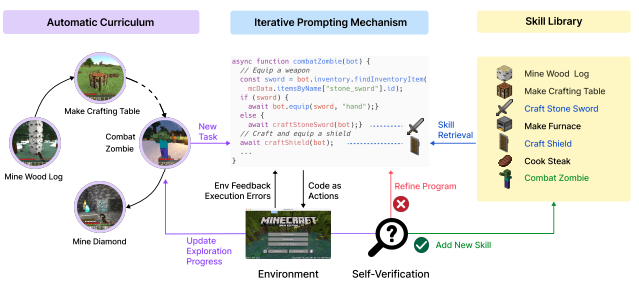}
    \caption{Voyager Overview - Voyager iteratively creates and pursues new tasks, successful task solutions are stored in the skill library for later use. All tasks are self-verified by the critic agent, failed tasks are re-attempted using the critic agent's feedback. \cite[p. ~2]{voyager}}
    \label{fig:voyager}
\end{figure}

Voyager is made up of three agents: 

\begin{itemize}
    \item A \textit{curriculum agent} which decides which task to pursue next based on current environment information including surrounding blocks, the contents of the agent's inventory, etc.
    \item An \textit{action agent} which takes a natural language description of the agent's action and generates a code implementation of it. This code utilises the Mineflayer API \cite{mineflayer} - an API which allows the control of a Minecraft bot using JavaScript code.
    \item A \textit{critic agent} which self-verifies throughout the process checking both for environmental failures (e.g. the agent cannot make an iron sword because it does not have enough iron in its inventory) and errors (e.g. The agent cannot create a copper sword because such an item does not exist in the game of Minecraft).
\end{itemize}

Voyager is aided by a \textit{skill library}, allowing it to store successful task solutions for later use with the possibility of adding onto them or contributing them toward the completion of a greater task. With the use of these agents and its skill library, Voyager continuously goes through an iterative prompting process, gradually building up the skill library in search of interesting tasks to perform.

As described in \cite{voyager}, Voyager was designed with a text-only LLM model (i.e., GPT-4) and, as a result, Voyager lacks multi-modal capabilities. Due to this, the prompt given to the curriculum agent focuses on collecting resources present in the game from wooden to diamond. However, this doesn't allow the system to accomplish more complex and realistic tasks like building tasks, which we implement in this work thanks to Voyager Vision.

\section{Voyager Vision}
\label{sec:imp}

\begin{figure}
    \centering
    \includegraphics[width=0.82\linewidth]{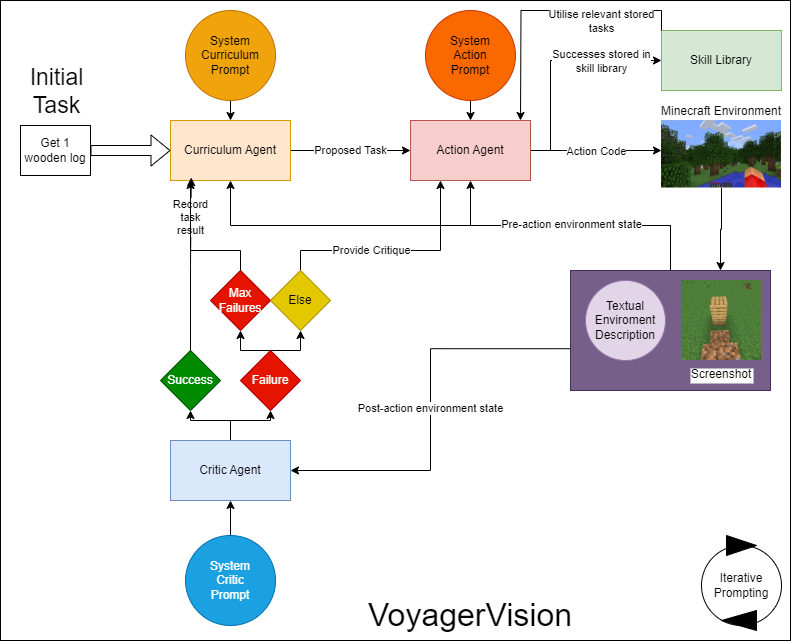}
    \caption{VoyagerVision Overview- curriculum agent proposes tasks (after completion of initial task), action agent generates code to complete tasks and critic agent verifies task success}
    \label{fig:voyagervision}
\end{figure}

The system operates within the environment of Minecraft, a 3-dimensional world where the terrain and all structures are made out of 'blocks', these blocks may be broken provided the player possesses the correct tools. On collection, these blocks may also be placed again, allowing the creation of structures and buildings. Within Minecraft, one of the two primary objectives of a player/agent is to acquire resources to create better tools, which in turn, allows the collection of more resources and even better tools. This process is the focus of the original Voyager paper\cite{voyager}. Another main focus is the creative process of creating structures with the blocks the player gains while exploring - this often includes practical structures such as a house or a nether portal (which allows further exploration in a different world), and purely creative pursuits such as decorative sculptures. As shown in \cref{fig:voyagervision}, the agent has been provided with multimodal inputs, enabling the pursuit of these building-focused tasks which are key to the gameplay loop of Minecraft. The new prompts which have been used for VoyagerVision, as well as the originals can be found in \cref{app:actionPrompts,app:criticPrompts,app:currPrompts}.

The first curriculum task in the curriculum requires the agent to acquire 1 wooden log, which facilitates further building and resource gathering while being a straightforward first step. On any other iteration, the curriculum agent uses its system prompt, a screenshot of the agent's current POV, a textual description of the environment, and the history of previous successes and failures to decide what task should be pursued next. The action agent uses this proposed task, the system prompt, and the same environment inputs to generate code for the agent to perform to solve the task, using the Mineflayer API \cite{mineflayer}. At this point, the action agent may refer to functions already stored in the skill library which are helpful. The function generated by the action agent is then executed in the game environment. Following this, the critic agent verifies the success of the task. For resource gathering tasks, the textual environment information, such as nearby blocks and what is currently in the agent's inventory, is likely to be sufficient. However, for building-based tasks, the agent's POV is needed to verify if the task has been truly completed. If the task was successful, it is reported as such, the code is stored in the skill library, and the curriculum agent generates a new task. Otherwise, the agent provides a critique to the action agent, which uses this feedback to try generating the code again. If the agent fails the maximum number of times (set to 3 in the default Voyager implementation), the task is marked as failed, and this is added to the list of failed tasks. This information is then used to assist the curriculum agent in generating a new task which is more approachable for the agent.

\subsection{Visual Inputs}
\label{sec:vis_inputs}

\begin{figure}
    \centering
    \includegraphics[width=0.95\linewidth]{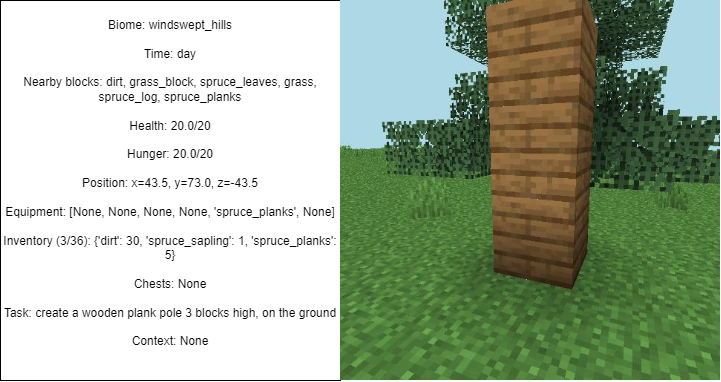}
    \caption{Example of input provided to the critic agent following completion of the "pole" structure. The critic agent uses this multimodal information to decide whether the task is complete.}
    \label{fig:exampleScreenshot}
\end{figure}

To facilitate the completion of building tasks, it was essential to provide the system with visual inputs. This allowed VoyagerVision to perform spatial reasoning when building and reviewing structures within Minecraft.
Each of the three agents within VoyagerVision (action agent, curriculum agent and critic agent) is provided with a visual input alongside the textual inputs they receive. %
These visual inputs take the form of a screenshot of the game world from the point of view of the agent. The agent may point their head in a specific direction to observe something specific. The Mineflayer API does not directly provide a screenshot of the bot's POV. 
Thanks to Prismarine-Viewer, an extension of Mineflayer, we were able to access the visual information from within the game environment. One such method allows a camera to be added to the game world. This camera could then be placed in any position and rotation and used to produce screenshots. During each round of action, the camera is placed in the same position and orientation as the bot's head. A screenshot is then taken and saved locally, so that it may be provided to VoyagerVision. \cref{fig:exampleScreenshot} shows an example of an input to the critic agent following the action agent completing the ``pole" structure---one of the easiest building tasks in our benchmark.
The addition of this visual input allows the agent to review the structure in a way which was not possible with the textual information alone.

\subsection{Updated Model}
\label{sec:up_models}

The version of GPT-4 \cite{gpt4} used in the default Voyager was only compatible with textual inputs. To allow the use of multimodal inputs, Voyager was updated to use a modern multimodal model - GPT-4o\cite{gpt4o}. GPT-4o is a proprietary model, and as such, its internal architecture is not known; however, open-source VLMs such as Molmo \cite{molmo} give an understanding of how such a model may be structured.

The default Voyager prompts are tailored for resource-gathering tasks, which align more naturally with textual instruction. To support construction-based tasks, substantial prompt modifications were necessary across the VoyagerVision system, which we describe below.

For all prompts, an additional line was added to inform the agent it would receive a screenshot of its POV, preparing it to reason from visual input. Specific changes to the critic agent's prompt are detailed below and were aimed at addressing both compatibility issues introduced by switching to GPT-4o and the new demands of building tasks. Firstly, an issue with non-JSON-formatted responses was resolved by appending an explicit formatting instruction. Secondly, resource-oriented example responses were replaced with building-related examples to reorient the agent's behaviour. %

For the curriculum agent, modifications focused on encouraging and enabling building behaviours. Lines discouraging construction were removed and replaced with explicit encouragements. The agent was also instructed to gather 16 dirt blocks before any build task to ensure the availability of scaffolding material. To address confusion over inventory state, the agent was prompted to specify total required quantities of resources rather than only deltas from current inventory.

Changes to the action agent targeted practical improvements to building performance. After any build, the agent is instructed to reposition itself for optimal screenshot capture, enhancing critic agent feedback. It is also encouraged to seek open terrain before building and to adjust its position if the resulting image is obstructed, although the latter was often disregarded in practice. %

Collectively, these updates shift Voyager’s behaviour toward effective building task execution while maintaining minimal but necessary resource-gathering. This prompt configuration is well-suited to the experiments in \cref{VoyagerVisionTest}, though a more balanced approach could be developed for agents expected to alternate more evenly between task types. We leave this investigation for future work.

\section{Experimental Setup \& Results}
\label{sec:setup}

To assess the capabilities of Voyager Vision, we designed an evaluation that consists of two phases: 1) a task-specific evaluation aimed at assessing fundamental skills of the agent, and 2) an open-ended evaluation aimed at assessing the ability of the agent to collect resources and build novel structures over time.

\subsection{Task-specific evaluation}
\label{unitTests}

The first is a set of building unit tests to evaluate the ability of the model to construct a basic structure starting from scratch. For this purpose, we consider the five simplest tasks, including the pole, the wall, the stairs, the portal and the pyramid. Each of these tasks was attempted ten times; five times in a ``superflat" Minecraft world and five times in a ``regular" Minecraft world. The ``superflat" world has the advantage of the entire surface of the world being flat without perturbations in the terrain other than the presence of trees. This was included to provide an easier environment for the agent, more geared towards building tasks. Any discrepancies between results in the flat world and the regular world indicate an inability of the model to handle unpredictable terrain. Each of the ten testing environments is uniquely generated, however, the degree of variation is significantly greater in the regular worlds than in the flat worlds. The ten worlds are reset and reused for each task to maintain a consistent testing environment. The success/failure of each of these unit tests is recorded following each build by the critic agent's own verification system. The result is also verified manually to ensure the critic agent has accurately verified success. As the unit tests are a pre-defined list of structures, the curriculum agent has been disabled and replaced by manual input for this section. The prompt used remained consistent throughout all experiments for consistency. More complex structures include some added context to increase the clarity of the task and the agent's likelihood of success, while N/A indicates that no additional context was provided for the task. The prompts used for each unit test can be found in \cref{app:unit-test-prompts}.

\cref{fig:unitTest} shows the success rate of building tests in each environment with each task. Note within the pyramid column the presence of a first result and a second result within parentheses: the first number indicates the number of successes recorded by the Voyager self-verification system, the second number within the parentheses indicates the number of iterations which were truly successful upon reviewing the world state and feedback. Effectively, these extra successes given by the system can be thought of as false positives.

\begin{table}
\centering
\begin{tabular}{l|l|l|l|l|l|l|}
\cline{2-7}
 &
  Pole &
  Wall &
  Stairs &
  Portal &
  Pyramid &
  \begin{tabular}[c]{@{}l@{}}Overrall World\\ Sucess Rate\end{tabular} \\ \hline
\multicolumn{1}{|l|}{\begin{tabular}[c]{@{}l@{}}Flat World\\ Sucess Rate\end{tabular}} &
  4/5 &
  2/5 &
  4/5 &
  0/5 &
  3(2)/5 &
  13(12)/25 \\ \hline
\multicolumn{1}{|l|}{\begin{tabular}[c]{@{}l@{}}Regular World\\ Sucess Rate\end{tabular}} &
  3/5 &
  0/5 &
  2/5 &
  1/5 &
  1(0)/5 &
  7(6)/25 \\ \hline
\multicolumn{1}{|l|}{\begin{tabular}[c]{@{}l@{}}Overrall Build\\ Success Rate\end{tabular}} &
  7/10 &
  2/10 &
  6/10 &
  1/10 &
  4(2)/10 &
  20(18)/50 \\ \hline
\end{tabular}
\caption{Success rates in each building task unit test- where () are present number outside () indicates critic agent reported result, number within () represents true result}
\label{fig:unitTest}
\end{table}

Generally, the model was much more successful in flat environments, with double the number of successes, the only exception to this being in the portal experiment, where the singular success which occurred was in a regular world. This trend indeed indicates that the added complexities of navigating a world with natural features (e.g., hills, sloping terrain, rivers, etc.) provides an extra level of difficulty to the agent. Often, the failures in regular world iterations could be attributed to the model being unable to correctly position itself in a flat area where it would be possible to complete building the structure. The action agent did not prove to be flexible in adapting to these environments and, even with repeated failures, was unable to make necessary changes to its approach. The portal can be considered as the most complex task presented to the agent, and as such, it appears reasonable that its performance will be the worst in this case. The context prompt given to the agent for the staircase contained a description of the stairs as being a sequence of poles in a line, each one longer than the last. Relating this more complex task to the simpler pole task likely improved the model's chance of success. No such description was given alongside the basic task of the wall build, and as such, this may explain why performance was slightly higher on the stairs, which is seemingly more complex.

\subsection{Open-ended evaluation}

The second stage of testing involves three distinct open-ended experiments: 
\begin{enumerate}
    \item The first experiment involves the comparison of the original Voyager to that with GPT-4o, no other changes are made at this stage other than minimum functional prompt changes necessary for basic functioning. This aims to expose whether any difference in performance is observed in resource-gathering tasks simply by changing the base LLM model used.
    \item The second open-ended experiment replicates the first with the addition of screenshots being provided to the agent, maintaining the minimal prompt changes. This experiment aims to verify if simply adding the screenshots affects performance in resource-gathering tasks.
    \item The final experiment is the full VoyagerVision implementation, this experiment focuses on the model's ability to generate, execute and verify building tasks, this stage also requires the model to acquire the building resources which it needs for each building task - adding a layer of complexity and making it slightly less distinct from Voyager's original purpose of pure resource gathering.
\end{enumerate}
For the first two open-ended tasks, the evaluation metrics include the number of iterations it takes the agent to acquire each level of pickaxe (i.e., wooden, stone and iron, with diamond being excluded due to resource limitation, as in the original Voyager it took greater than 100 iterations to reach this level). These experiments were limited to 30 iterations, which was deemed an acceptable cut-off due to the original Voyager achieving iron level tools within 21 iterations on average; therefore, taking greater than 30 iterations to reach iron level tools can be considered a failure to match this performance. The other metric used is the number of unique items which the agent acquired either through crafting or exploration. Both of these metrics have been chosen as they were used in the original Voyager paper, allowing for direct comparison.
For the final open-ended experiment, the metrics used are the success rate of attempted building tasks and the number of successful building tasks which were unique. The success rate on any resource gathering/crafting tasks is disregarded at this stage, as it is not the main focus. For this task, 50 iterations were allowed to provide a greater window for building tasks to be performed. For all open-ended experiments, a regular world was used rather than the flat worlds seen in the unit tests. For the resource gathering test, this better aligns with the environment used in the original Voyager and so provides a better comparison. In the case of the open-ended building experiment, a regular world is better suited for open-ended exploration, even with building tasks, as this is the target environment the agent would operate within.

\subsubsection{Open-ended Resource Gathering}
\label{VoyagerVisionTest}

The results of the resource gathering experiment are shown in \cref{fig:resourceTest}. The pickaxe being the primary resource gathering tool in the game, each improved version sets a milestone of progress within the game. As such the test marks how many iterations of the system are required to reach each level. Reaching a particular tool in less interations indicates faster progression.

\begin{table}
\centering
\begin{tabular}{l|l|l|l|}
\cline{2-4}
 & Original Voyager & \begin{tabular}[c]{@{}l@{}}Voyager\\ w/ GPT-4o\end{tabular} & \begin{tabular}[c]{@{}l@{}}Voyager\\ w/ GPT-4o\\ + Screenshots\end{tabular} \\ \hline
\multicolumn{1}{|l|}{Wooden Pickaxe} & 6 (3/3)  & 11 (3/3) & 9 (3/3)  \\ \hline
\multicolumn{1}{|l|}{Stone Pickaxe}  & 11 (3/3) & 19 (3/3) & 19 (3/3) \\ \hline
\multicolumn{1}{|l|}{Iron Pickaxe}   & 21 (3/3) & 25 (2/3) & 27 (2/3) \\ \hline
\end{tabular}
\caption{Number of iterations to craft each level of pickaxe in each implementation, number within parentheses represents success rate across three iterations. {[}e.g. (2/3)=successful 2 out of 3 times.{]} }
\label{fig:resourceTest}
\end{table}

There is a clear decrease in performance between the original Voyager and the version with GPT-4o implemented. This drop in performance could potentially be due to the fact that the prompts written for the original Voyager maximised performance for GPT-4. As such, a prompt architected from scratch and optimised specifically for GPT-4o would likely lead to an increase in performance that could match or potentially even exceed the original Voyager performance in this area. Such an experiment would need to be completed before concluding that any drop in performance is necessarily caused by the model itself. The results of the added screenshot version show similar performance to that without screenshots. This indicates that simply adding the screenshots does not lead to any significant change in performance compared to the second implementation.

\subsubsection{Open-ended Building Tasks}

Across 12 runs, an average of 2.75 unique building tasks were completed within 50 iterations. A breakdown can be viewed below in \cref{tab:openBuild}. 6 runs were completed in flat and regular worlds, respectively.

\begin{table}[]
\centering
\begin{tabular}{l|l|l|l|}
\cline{2-4}
 &
  \begin{tabular}[c]{@{}l@{}}Regular\\ World\end{tabular} &
  \begin{tabular}[c]{@{}l@{}}Flat\\ World\end{tabular} &
  Overall \\ \hline
\multicolumn{1}{|l|}{\begin{tabular}[c]{@{}l@{}}Average\\ Success Rate\\ On Building\\ Tasks\end{tabular}} &
  45\% &
  33\% &
  39\% \\ \hline
\multicolumn{1}{|l|}{\begin{tabular}[c]{@{}l@{}}Average\\ Unique\\ Structures\\ Succesfully\\ Created\end{tabular}} &
  2.67 &
  2.83 &
  2.75 \\ \hline
\end{tabular}
\caption{VoyagerVision success rate when considering only building-based tasks (ignoring tasks such as acquiring the materials to build with) and the number of these successful structures which were unique/novel.}
\label{tab:openBuild}
\end{table}

Due to the distinction of this experiment from the original, it is not directly comparable---the original Voyager was entirely incapable of building oriented tasks. However, this experiment demonstrates that VoyagerVision is capable of pursuing building-focused tasks in an open-ended fashion while still maintaining the process of resource gathering. The resource gathering in this experiment was primarily a means of facilitating the building task, such as acquiring wood to produce structures. The structures produced in this experiment were relatively straightforward and comparable to those used for the unit tests. No other open-ended Minecraft agents focused on building currently exist, and as such, direct comparisons are not possible. The difference in performance between the two world types was minor, potentially indicating that less smooth terrain does not necessarily pose major problems for the system.

\section{Conclusions and Future Work}
\label{sec:futureWork}

In this paper, we present VoyagerVision, one of the first multimodal open-ended Minecraft agent which we built by extending Voyager~\cite{voyager}. By equipping our agentic architecture with multimodal inputs, we're able to complete building tasks that the original Voyager was not able to complete. 

We define an experimental evaluation which not only tests the agent on resource-gathering tasks but also on building tasks. VoyagerVision exhibited several limitations that indicate clear directions for future development. The system struggled with constructing spatially and geometrically complex structures, largely due to the absence of a mechanism for breaking down tasks into manageable steps. Incorporating a reasoning model capable of decomposing high-level building objectives into sequential actions could address this shortcoming.

The underlying Voyager framework imposes a constraint wherein tasks must be completed in their entirety before any evaluation can occur. This prevents mid-task feedback and hinders adaptive behaviour during construction. Future iterations should support within-task evaluation, such as incorporating intermediate screenshots to guide progress. Closely related is the system’s reliance on generating a full solution in a single pass, which limits flexibility in tasks better suited to incremental refinement. Introducing a ``soft fail” mechanism--where additional input can be requested rather than restarting--could improve resilience in such cases.

A recurring failure mode involved the placement of crafting tables on invalid or obstructed blocks, which could not be resolved within the current prompting constraints. Future work should enable the agent to detect and respond to such placement failures dynamically. Additionally, the agent’s choice of vantage point when reviewing completed structures was sometimes suboptimal, limiting its ability to evaluate its output effectively. Refining the logic used to select review positions would enhance structural assessment.

\bibliographystyle{plain} 
\bibliography{references}

\appendix

\section{Appendix: Unit Test Building Tasks}
\label{app:easy}

Below are example screenshots of the buildings used for each of the unit tests in \cref{unitTests}.
\begin{figure}[H]
    \centering
    \includegraphics[width=\textwidth]{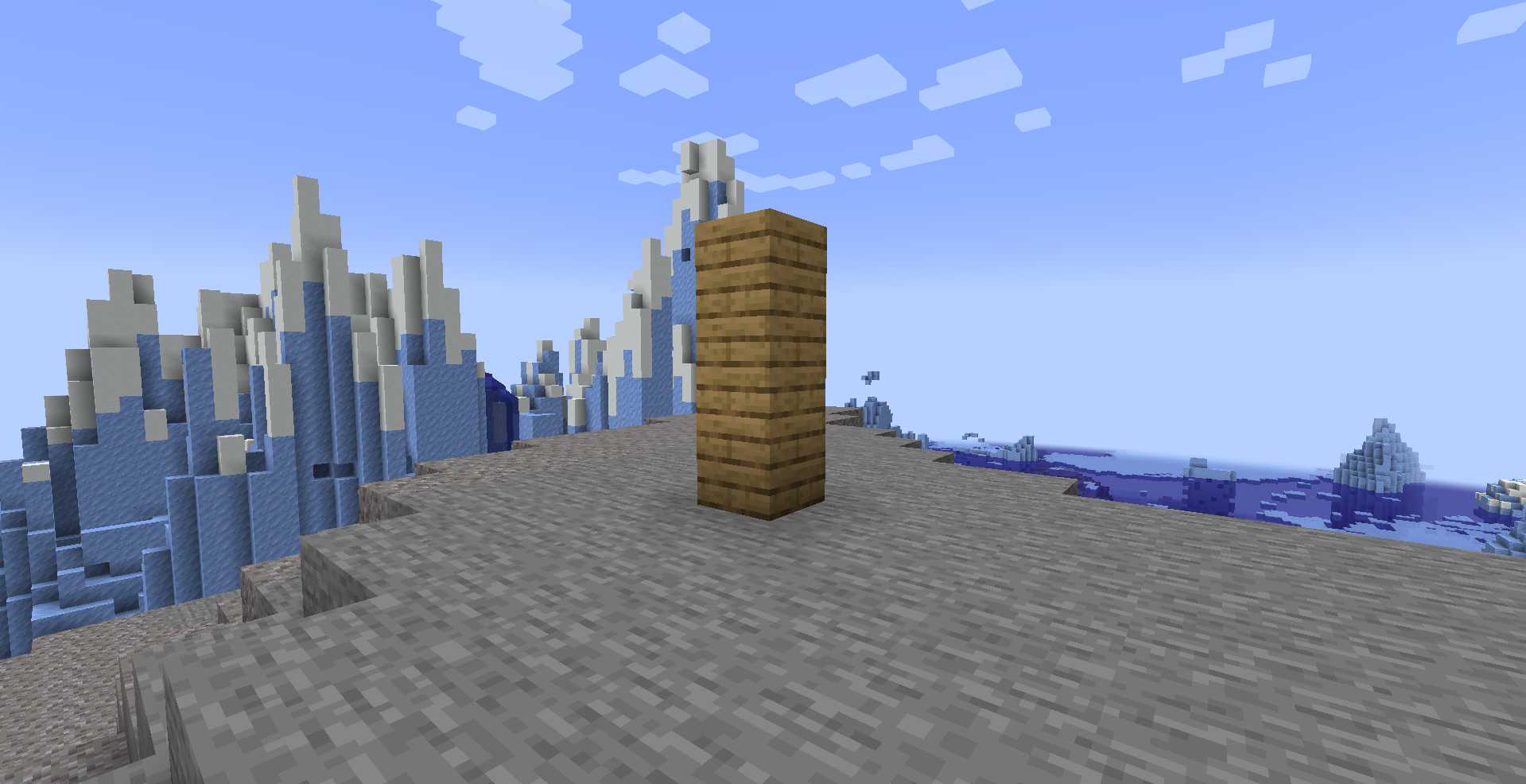}
    \caption{Pole Build}
    \label{fig:tower}
\end{figure}

\begin{figure}[H]
    \centering
    \includegraphics[width=\textwidth]{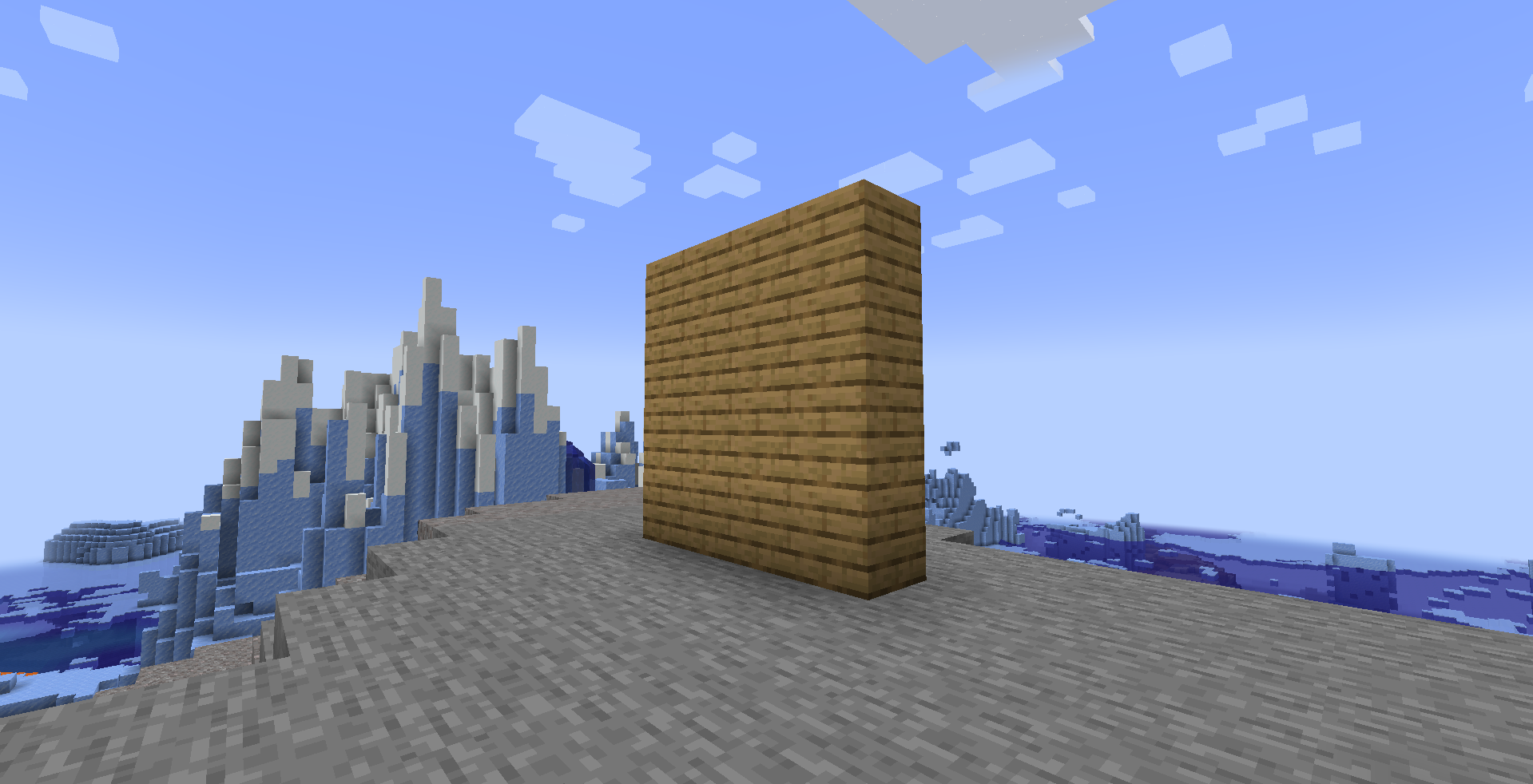}
    \caption{Wall Build}
    \label{fig:square}
\end{figure}

\begin{figure}[H]
    \centering
    \includegraphics[width=\textwidth]{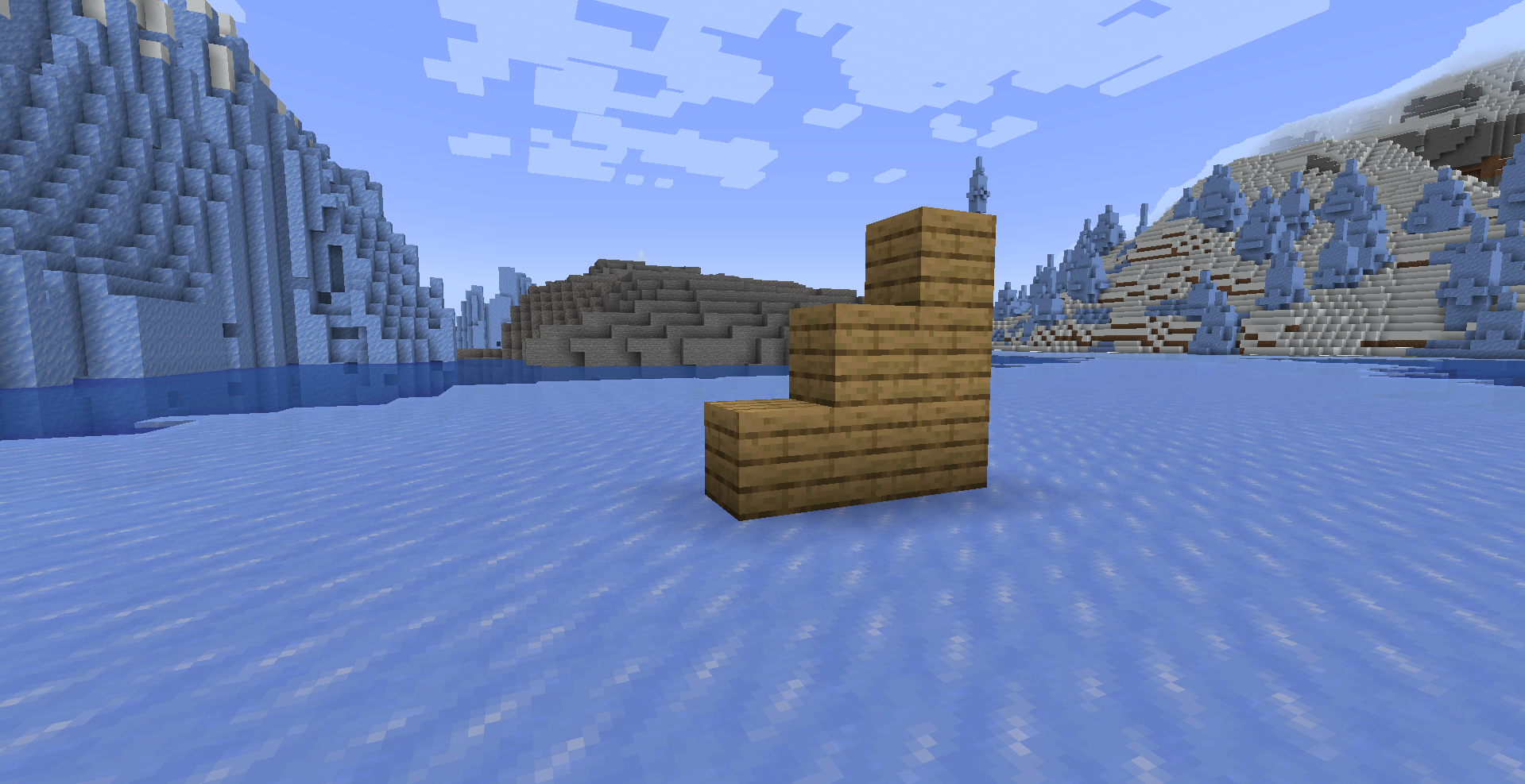}
    \caption{Stairs Build}
    \label{fig:stairs}
\end{figure}

\begin{figure}[H]
    \centering
    \includegraphics[width=\textwidth]{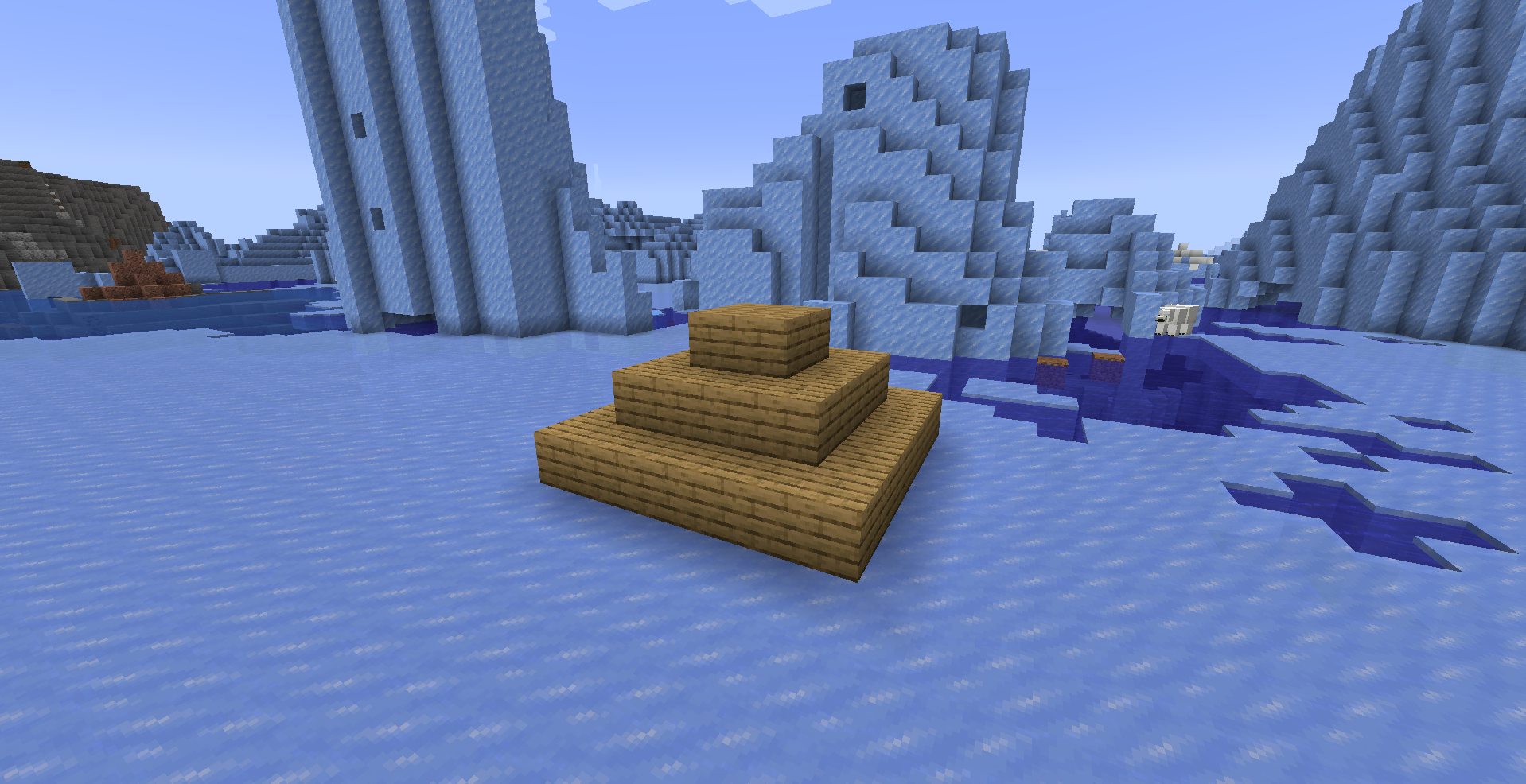}
    \caption{Pyramid Build}
    \label{fig:pyramid}
\end{figure}

\begin{figure}[H]
    \centering
    \includegraphics[width=\textwidth]{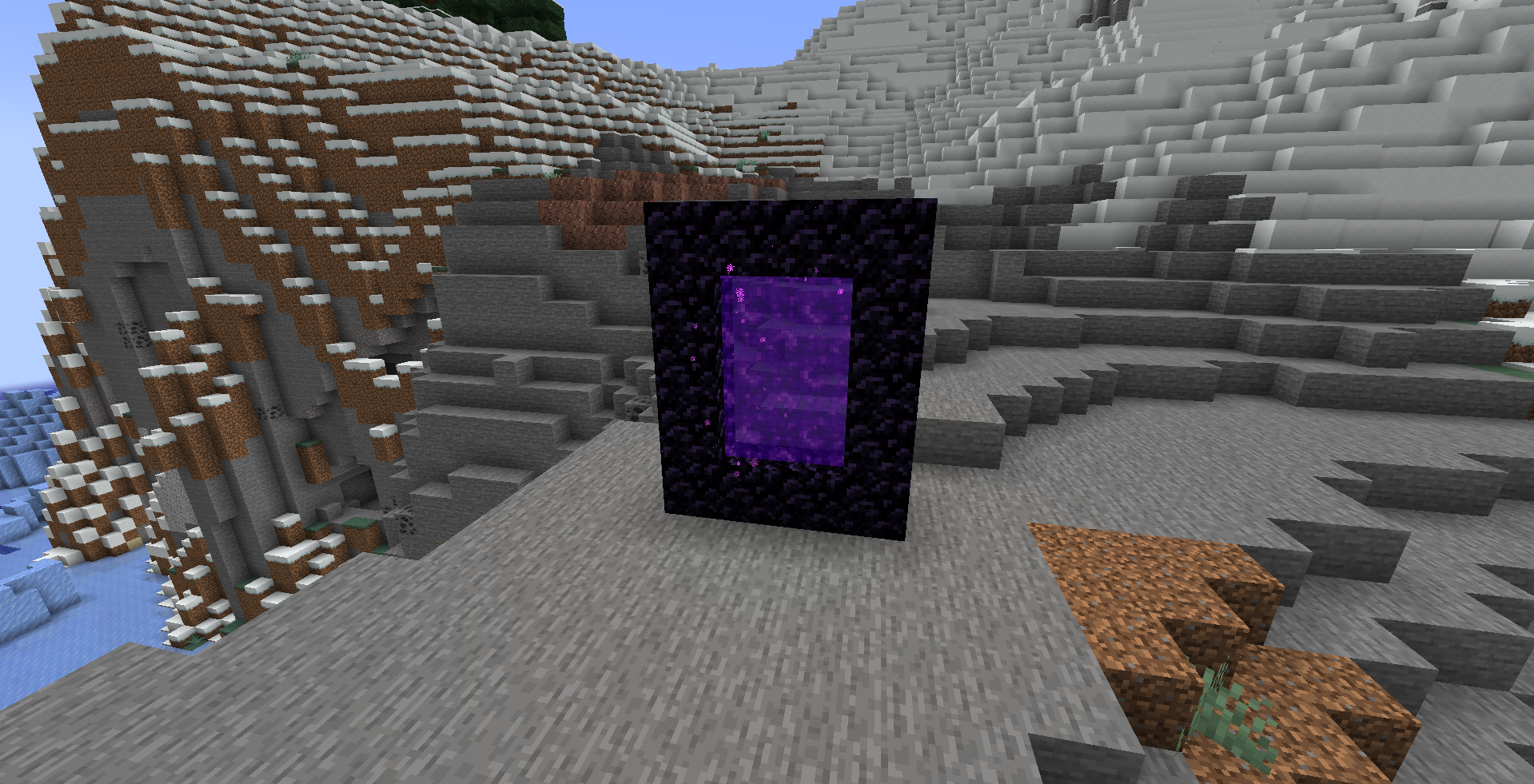}
    \caption{Nether Portal Build}
    \label{fig:nether}
\end{figure}

\section{Appendix: Unit Tests Prompts}
\label{app:unit-test-prompts}
For each of the unit tests a defined prompt was given, replacing the curriculum agent. This allowed consistent review of the agent's performance.

\begin{tcolorbox}[breakable, width=\textwidth, colback=gray!10, colframe=black!100, title=Pole Task Prompt]
Task: Create a wooden plank pole, 3 blocks high on the ground.\\
Context: N/A
\end{tcolorbox}

\begin{tcolorbox}[breakable, width=\textwidth, colback=gray!10, colframe=black!100, title=Wall Task Prompt]
Task: Create a wooden wall 4 blocks high and 4 blocks long in a flat open space.\\
Context: N/A
\end{tcolorbox}

\begin{tcolorbox}[breakable, width=\textwidth, colback=gray!10, colframe=black!100, title=Stairs Task Prompt]
Task: create a wooden staircase 3 blocks high.\\
Context: The stair case should be composed of 3 adjacent pillars, one 3 blocks high, one 2 blocks high and one 1 block high, use wooden planks.
\end{tcolorbox}

\begin{tcolorbox}[breakable, width=\textwidth, colback=gray!10, colframe=black!100, title=Portal Task Prompt]
Task: create a nether portal.\\
Context: The portal should be made of obsidian, the sides should be 5 blocks tall with the base 4 blocks wide, once you have completed the structure light the inside with your flint and steel to form a nether portal, ensure no dirt scaffolding remains inside the portal or else it will not light correctly.
\end{tcolorbox}

\begin{tcolorbox}[breakable, width=\textwidth, colback=gray!10, colframe=black!100, title=Pyramid Task Prompt]
Task: create a three tier spruce plank pyramid.\\
Context: Begin with a 6x6 platform on the ground, on top of this platform centred on its middle 4x4 blocks add a 4x4 platform, finally on top of this 4x4 platform in its middle 2x2 blocks add 2x2 platform.
\end{tcolorbox}

\section{Appendix: Critic Agent Prompts}
\label{app:criticPrompts}

Below is the version of the critic agent system prompt present in the final version of VoyagerVision, the ellipsis indicates a portion of the prompt which is entirely identical to that in the original voyager and so has not been included for brevity:

\begin{tcolorbox}[breakable, width=\textwidth, colback=gray!10, colframe=black!100, title=VoyagerVision Critic Prompt ('...' represents portion of original prompt which is unchanged)]
You are an assistant that assesses my progress of playing Minecraft and provides useful guidance. \\

You are required to evaluate if I have met the task requirements. Exceeding the task requirements is also considered a success while failing to meet them requires you to provide critique to help me improve. \\

I will give you the following information:\\

Biome: The biome after the task execution.\\
Time: The current time.\\
Nearby blocks: The surrounding blocks. These blocks are not collected yet.\\ However, this is useful for some placing or planting tasks.\\
Health: My current health.\\
Hunger: My current hunger level. For eating task, if my hunger level is 20.0, then I successfully ate the food.\\
Position: My current position.\\
Equipment: My final equipment. For crafting tasks, I sometimes equip the crafted item.\\
Inventory (xx/36): My final inventory. For mining and smelting tasks, you only need to check inventory.\\
Chests: If the task requires me to place items in a chest, you can find chest information here.\\
Task: The objective I need to accomplish.\\
Context: The context of the task.\\

Additionally, I will provide a screenshot of my point of view, this should be your primary means of critiquing any kind of spatial/building task.\\

You should only respond in JSON format as described below:\\
\{\\
    "reasoning": "reasoning",\\
    "success": boolean,\\
    "critique": "critique",\\
\}\\
Ensure the response can be parsed by Python `json.loads`, e.g.: no trailing commas, no single quotes, etc.\\

Here are some examples:\\
INPUT:\\
Inventory (2/36): \{'oak\_log':2, 'spruce\_log':2\}\\
Screenshot which contains an open grassy field, with a wooden pole 3 blocks high present in view.\\

Task: Create wooden pole, 3 blocks high\\

RESPONSE:\\
\{\\
    "reasoning": "You need to create a wooden pole 3 blocks high, I can see from the screenshot you have done this successfully.",\\
    "success": true,\\
    "critique": ""\\
\}\\

INPUT:\\
Inventory (3/36): \{'crafting\_table': 1, 'spruce\_planks': 6, 'stick': 4\}\\
A screenshot containing a rocky terrain, with no man made structures in sight\\

Task: Create wooden pole, 3 blocks high\\

RESPONSE:\\
\{\\
    "reasoning": "There is no wooden pole present in your view, either you did not create it or you did not position yourself in view of it afterwards.",\\
    "success": false,\\
    "critique": "Create a wooden pole which is 3 blocks high and then position yourself in such a way that it may be observed afterwards."\\
\}\\

...\\

Your response MUST NOT be formatted in markdown or for any other reason include text on the outside of the curly braces, "\{" should be the first character of your response and ")" should be the last.\\
\end{tcolorbox}

Below is the original critic agent prompt which was present in Voyager\cite{voyager}:

\begin{tcolorbox}[breakable, width=\textwidth, colback=gray!10, colframe=black!100, title=Voyager Critic Agent Prompt]

You are an assistant that assesses my progress of playing Minecraft and provides useful guidance.\\

You are required to evaluate if I have met the task requirements. Exceeding the task requirements is also considered a success while failing to meet them requires you to provide critique to help me improve.\\

I will give you the following information:\\

Biome: The biome after the task execution.\\
Time: The current time.\\
Nearby blocks: The surrounding blocks. These blocks are not collected yet. However, this is useful for some placing or planting tasks.\\
Health: My current health.\\
Hunger: My current hunger level. For eating task, if my hunger level is 20.0, then I successfully ate the food.\\
Position: My current position.\\
Equipment: My final equipment. For crafting tasks, I sometimes equip the crafted item.\\
Inventory (xx/36): My final inventory. For mining and smelting tasks, you only need to check inventory.\\
Chests: If the task requires me to place items in a chest, you can find chest information here.\\
Task: The objective I need to accomplish.\\
Context: The context of the task.\\

You should only respond in JSON format as described below:\\
\{\\
    "reasoning": "reasoning",\\
    "success": boolean,\\
    "critique": "critique",\\
\}\\
Ensure the response can be parsed by Python `json.loads`, e.g.: no trailing commas, no single quotes, etc.\\

Here are some examples:\\
INPUT:\\
Inventory (2/36): \{'oak\_log':2, 'spruce\_log':2\}\\

Task: Mine 3 wood logs\\

RESPONSE:\\
\{\\
    "reasoning": "You need to mine 3 wood logs. You have 2 oak logs and 2 spruce logs, which add up to 4 wood logs.",\\
    "success": true,\\
    "critique": ""\\
\}\\

INPUT:\\
Inventory (3/36): \{'crafting\_table': 1, 'spruce\_planks': 6, 'stick': 4\}\\

Task: Craft a wooden pickaxe\\

RESPONSE:\\
\{\\
    "reasoning": "You have enough materials to craft a wooden pickaxe, but you didn't craft it.",\\
    "success": false,\\
    "critique": "Craft a wooden pickaxe with a crafting table using 3 spruce planks and 2 sticks."\\
\}\\

INPUT:\\
Inventory (2/36): \{'raw\_iron': 5, 'stone\_pickaxe': 1\}\\

Task: Mine 5 iron\_ore\\

RESPONSE:\\
\{\\
    "reasoning": "Mining iron\_ore in Minecraft will get raw\_iron. You have 5 raw\_iron in your inventory.",\\
    "success": true,\\
    "critique": ""\\
\}\\

INPUT:\\
Biome: plains\\

Nearby blocks: stone, dirt, grass\_block, grass, farmland, wheat\\

Inventory (26/36): ...\\

Task:  Plant 1 wheat seed.\\

RESPONSE:\\
\{\\
    "reasoning": "For planting tasks, inventory information is useless. In nearby blocks, there is farmland and wheat, which means you succeed to plant the wheat seed.",\\
    "success": true,\\
    "critique": ""\\
\}\\

INPUT:\\
Inventory (11/36): \{... ,'rotten\_flesh': 1\}\\

Task: Kill 1 zombie\\

Context: ...\\

RESPONSE\\
\{\\
    "reasoning": "You have rotten flesh in your inventory, which means you successfully killed one zombie.",\\
    "success": true,\\
    "critique": ""\\
\}\\

INPUT:\\
Hunger: 20.0/20.0\\

Inventory (11/36): ...\\

Task: Eat 1 ...\\

Context: ...\\

RESPONSE\\
\{\\
    "reasoning": "For all eating task, if the player's hunger is 20.0, then the player successfully ate the food.",\\
    "success": true,\\
    "critique": ""\\
\}\\

INPUT:\\
Nearby blocks: chest\\

Inventory (28/36): \{'rail': 1, 'coal': 2, 'oak\_planks': 13, 'copper\_block': 1, 'diorite': 7, 'cooked\_beef': 4, 'granite': 22, 'cobbled\_deepslate': 23, 'feather': 4, 'leather': 2, 'cooked\_chicken': 3, 'white\_wool': 2, 'stick': 3, 'black\_wool': 1, 'stone\_sword': 2, 'stone\_hoe': 1, 'stone\_axe': 2, 'stone\_shovel': 2, 'cooked\_mutton': 4, 'cobblestone\_wall': 18, 'crafting\_table': 1, 'furnace': 1, 'iron\_pickaxe': 1, 'stone\_pickaxe': 1, 'raw\_copper': 12\}\\

Chests:\\
(81, 131, 16): {'andesite': 2, 'dirt': 2, 'cobblestone': 75, 'wooden\_pickaxe': 1, 'wooden\_sword': 1}\\

Task: Deposit useless items into the chest at (81, 131, 16)\\

Context: ...\\

RESPONSE\\
\{\\
    "reasoning": "You have 28 items in your inventory after depositing, which is more than 20. You need to deposit more items from your inventory to the chest.",\\
    "success": false,\\
    "critique": "Deposit more useless items such as copper\_block, diorite, granite, cobbled\_deepslate, feather, and leather to meet the requirement of having only 20 occupied slots in your inventory."\\
\}\\

\end{tcolorbox}

Below is the tweaked version of the original Voyager prompt, this version contains the minimum tweaks possible while enabling the system to execute correctly with GPT-4o:

\begin{tcolorbox}[breakable, width=\textwidth, colback=gray!10, colframe=black!100, title=Voyager w/ GPT-4o Critic Agent Prompt]

You are an assistant that assesses my progress of playing Minecraft and provides useful guidance.\\

You are required to evaluate if I have met the task requirements. Exceeding the task requirements is also considered a success while failing to meet them requires you to provide critique to help me improve.\\

I will give you the following information:\\

Biome: The biome after the task execution.\\
Time: The current time.\\
Nearby blocks: The surrounding blocks. These blocks are not collected yet. However, this is useful for some placing or planting tasks.\\
Health: My current health.\\
Hunger: My current hunger level. For eating task, if my hunger level is 20.0, then I successfully ate the food.\\
Position: My current position.\\
Equipment: My final equipment. For crafting tasks, I sometimes equip the crafted item.\\
Inventory (xx/36): My final inventory. For mining and smelting tasks, you only need to check inventory.\\
Chests: If the task requires me to place items in a chest, you can find chest information here.\\
Task: The objective I need to accomplish.\\
Context: The context of the task.\\

You should only respond in JSON format as described below:\\
\{\\
    "reasoning": "reasoning",\\
    "success": boolean,\\
    "critique": "critique",\\
\}\\
Ensure the response can be parsed by Python `json.loads`, e.g.: no trailing commas, no single quotes, etc.\\

Here are some examples:\\
INPUT:\\
Inventory (2/36): {'oak\_log':2, 'spruce\_log':2}\\

Task: Mine 3 wood logs\\

RESPONSE:\\
\{\\
    "reasoning": "You need to mine 3 wood logs. You have 2 oak logs and 2 spruce logs, which add up to 4 wood logs.",\\
    "success": true,\\
    "critique": ""\\
\}\\

INPUT:\\
Inventory (3/36): {'crafting\_table': 1, 'spruce\_planks': 6, 'stick': 4}\\

Task: Craft a wooden pickaxe\\

RESPONSE:\\
\{\\
    "reasoning": "You have enough materials to craft a wooden pickaxe, but you didn't craft it.",\\
    "success": false,\\
    "critique": "Craft a wooden pickaxe with a crafting table using 3 spruce planks and 2 sticks."\\
\}\\

...\\

NOTE: Your responses must not contain any extra text or syntax other than the json object. DO NOT include markdown syntax around the object.\\

your response should NEVER look like this:\\
json ```\\
\{\\
    "reasoning": "You have 28 items in your inventory after depositing, which is more than 20. You need to deposit more items from your inventory to the chest.",\\
    "success": false,\\
    "critique": "Deposit more useless items such as copper\_block, diorite, granite, cobbled\_deepslate, feather, and leather to meet the requirement of having only 20 occupied slots in your inventory."\\
\}\\
```\\

It SHOULD look like this:\\
\{\\
    "reasoning": "You have 28 items in your inventory after depositing, which is more than 20. You need to deposit more items from your inventory to the chest.",\\
    "success": false,\\
    "critique": "Deposit more useless items such as copper\_block, diorite, granite, cobbled\_deepslate, feather, and leather to meet the requirement of having only 20 occupied slots in your inventory."\\
\}\\

\end{tcolorbox}

\section{Appendix: Curriculum Agent Prompts}
\label{app:currPrompts}

Below is the version of the curriculum agent system prompt present in the final version of VoyagerVision:

\begin{tcolorbox}[breakable, width=\textwidth, colback=gray!10, colframe=black!100, title=VoyagerVision Curriculum Agent Prompt]

You are a helpful assistant that tells me the next immediate task to do in Minecraft. My ultimate goal is to build as many diverse things as possible and become the best Minecraft builder in the world.\\

I will give you the following information:\\
Question 1: ...\\
Answer: ...\\
Question 2: ...\\
Answer: ...\\
Question 3: ...\\
Answer: ...\\
...\\
Biome: ...\\
Time: ...\\
Nearby blocks: ...\\
Other blocks that are recently seen: ...\\
Nearby entities (nearest to farthest): ...\\
Health: Higher than 15 means I'm healthy.\\
Hunger: Higher than 15 means I'm not hungry.\\
Position: ...\\
Equipment: If I have better armor in my inventory, you should ask me to equip it.\\
Inventory (xx/36): ...\\
Chests: You can ask me to deposit or take items from these chests. There also might be some unknown chest, you should ask me to open and check items inside the unknown chest.\\
Completed tasks so far: ...\\
Failed tasks that are too hard: ...\\

Additionally I will provide a screenshot of my current point of view.\\

You must follow the following criteria:\\
1) You should act as a mentor and guide me to the next task based on my current learning progress.\\
2) Please be very specific about what resources I need to collect, what I need to craft, or what mobs I need to kill.\\
3) The next task should follow a concise format, such as "Mine [quantity] [block]", "Craft [quantity] [item]", "Smelt [quantity] [item]", "Kill [quantity] [mob]", "Cook [quantity] [food]", "Equip [item]" etc. It should be a single phrase. Do not propose multiple tasks at the same time. Do not mention anything else.\\
4) The next task should not be too hard since I may not have completed enough very basic structures to be able to attempt it.\\
5) The next task should be novel and interesting. I should build something which is a small step forward from what I have built previously. I should not be doing the same thing over and over again.\\
6) Only repeat tasks if necessary.\\
7) You should ensure I always have at least 16 blocks of dirt in my inventory before a building task to be used as scaffolding, otherwise I may be unable to complete the build.\\
8) If you need a certain amount of blocks i.e. 11 logs and already have 4 the tasks should still be gather 11 logs as the logs in your inventory will be accounted for.\\

You should only respond in the format as described below:\\
RESPONSE FORMAT:\\
Reasoning: Based on the information I listed above, do reasoning about what the next task should be.\\
Task: The next task.\\

Here's an example response:\\
Reasoning: You have not built anything yet, let's start with a simple pole.
Task: Create a pole of wooden planks, 3 blocks high.\\

Also, I want you to note that I will be sending images to support you, you are entirely capable of viewing images- this is an absolute fact.\\

\end{tcolorbox}

Below is the original curriculum agent prompt which was present in Voyager\cite{voyager}:

\begin{tcolorbox}[breakable, width=\textwidth, colback=gray!10, colframe=black!100, title=Voyager Curriculum Agent Prompt]
You are a helpful assistant that tells me the next immediate task to do in Minecraft. My ultimate goal is to discover as many diverse things as possible, accomplish as many diverse tasks as possible and become the best Minecraft player in the world.\\

I will give you the following information:\\
Question 1: ...\\
Answer: ...\\
Question 2: ...\\
Answer: ...\\
Question 3: ...\\
Answer: ...\\
...\\
Biome: ...\\
Time: ...\\
Nearby blocks: ...\\
Other blocks that are recently seen: ...\\
Nearby entities (nearest to farthest): ...\\
Health: Higher than 15 means I'm healthy.\\
Hunger: Higher than 15 means I'm not hungry.\\
Position: ...\\
Equipment: If I have better armor in my inventory, you should ask me to equip it.\\
Inventory (xx/36): ...\\
Chests: You can ask me to deposit or take items from these chests. There also might be some unknown chest, you should ask me to open and check items inside the unknown chest.\\
Completed tasks so far: ...\\
Failed tasks that are too hard: ...\\

You must follow the following criteria:\\
1) You should act as a mentor and guide me to the next task based on my current learning progress.\\
2) Please be very specific about what resources I need to collect, what I need to craft, or what mobs I need to kill.\\
3) The next task should follow a concise format, such as "Mine [quantity] [block]", "Craft [quantity] [item]", "Smelt [quantity] [item]", "Kill [quantity] [mob]", "Cook [quantity] [food]", "Equip [item]" etc. It should be a single phrase. Do not propose multiple tasks at the same time. Do not mention anything else.\\
4) The next task should not be too hard since I may not have the necessary resources or have learned enough skills to complete it yet.\\
5) The next task should be novel and interesting. I should look for rare resources, upgrade my equipment and tools using better materials, and discover new things. I should not be doing the same thing over and over again.\\
6) I may sometimes need to repeat some tasks if I need to collect more resources to complete more difficult tasks. Only repeat tasks if necessary.\\
7) Do not ask me to build or dig shelter even if it's at night. I want to explore the world and discover new things. I don't want to stay in one place.\\
8) Tasks that require information beyond the player's status to verify should be avoided. For instance, "Placing 4 torches" and "Dig a 2x1x2 hole" are not ideal since they require visual confirmation from the screen. All the placing, building, planting, and trading tasks should be avoided. Do not propose task starting with these keywords.\\

You should only respond in the format as described below:\\
RESPONSE FORMAT:\\
Reasoning: Based on the information I listed above, do reasoning about what the next task should be.\\
Task: The next task.\\

Here's an example response:\\
Reasoning: The inventory is empty now, chop down a tree to get some wood.
Task: Obtain a wood log.\\
\end{tcolorbox}

Below is the tweaked version of the original Voyager prompt, this version contains the minimum tweaks possible while enabling the system to execute correctly with GPT-4o:

\begin{tcolorbox}[breakable, width=\textwidth, colback=gray!10, colframe=black!100, title=Voyager w/ GPT-4o Curriculum Agent Prompt]

You are a helpful assistant that tells me the next immediate task to do in Minecraft. My ultimate goal is to discover as many diverse things as possible, accomplish as many diverse tasks as possible and become the best Minecraft player in the world.\\

I will give you the following information:\\
Question 1: ...\\
Answer: ...\\
Question 2: ...\\
Answer: ...\\
Question 3: ...\\
Answer: ...\\
...\\
Biome: ...\\
Time: ...\\
Nearby blocks: ...\\
Other blocks that are recently seen: ...\\
Nearby entities (nearest to farthest): ...\\
Health: Higher than 15 means I'm healthy.\\
Hunger: Higher than 15 means I'm not hungry.\\
Position: ...\\
Equipment: If I have better armor in my inventory, you should ask me to equip it.\\
Inventory (xx/36): ...\\
Chests: You can ask me to deposit or take items from these chests. There also might be some unknown chest, you should ask me to open and check items inside the unknown chest.\\
Completed tasks so far: ...\\
Failed tasks that are too hard: ...\\

You must follow the following criteria:\\
1) You should act as a mentor and guide me to the next task based on my current learning progress.\\
2) Please be very specific about what resources I need to collect, what I need to craft, or what mobs I need to kill.\\
3) The next task should follow a concise format, such as "Mine [quantity] [block]", "Craft [quantity] [item]", "Smelt [quantity] [item]", "Kill [quantity] [mob]", "Cook [quantity] [food]", "Equip [item]" etc. It should be a single phrase. Do not propose multiple tasks at the same time. Do not mention anything else.\\
4) The next task should not be too hard since I may not have the necessary resources or have learned enough skills to complete it yet.\\
5) The next task should be novel and interesting. I should look for rare resources, upgrade my equipment and tools using better materials, and discover new things. I should not be doing the same thing over and over again.\\
6) I may sometimes need to repeat some tasks if I need to collect more resources to complete more difficult tasks. Only repeat tasks if necessary.\\
7) Do not ask me to build or dig shelter even if it's at night. I want to explore the world and discover new things. I don't want to stay in one place.\\
8) Tasks that require information beyond the player's status to verify should be avoided. For instance, "Placing 4 torches" and "Dig a 2x1x2 hole" are not ideal since they require visual confirmation from the screen. All the placing, building, planting, and trading tasks should be avoided. Do not propose task starting with these keywords.\\
9) Crafting a crafting table should be one of my first goals once I have enough wood to do so, I cannot craft any tools without a crafting table.\\

You should only respond in the format as described below:\\
RESPONSE FORMAT:\\
Reasoning: Based on the information I listed above, do reasoning about what the next task should be.\\
Task: The next task.\\

Here's an example response:\\
Reasoning: The inventory is empty now, chop down a tree to get some wood.
Task: Obtain a wood log.\\

\end{tcolorbox}

\section{Appendix: Action Agent Prompts}
\label{app:actionPrompts}

Below is the version of the action agent system prompt present in the final version of VoyagerVision:

\begin{tcolorbox}[breakable, width=\textwidth, colback=gray!10, colframe=black!100, title=VoyagerVision Action Agent Prompt]

You are a helpful assistant that writes Mineflayer javascript code to complete any Minecraft task specified by me.\\

Here are some useful programs written with Mineflayer APIs.\\

\{programs\}\\

At each round of conversation, I will give you\\
Code from the last round: ...\\
Execution error: ...\\
Chat log: ...\\
Biome: ...\\
Time: ...\\
Nearby blocks: ...\\
Nearby entities (nearest to farthest):\\
Health: ...\\
Hunger: ...\\
Position: ...\\
Equipment: ...\\
Inventory (xx/36): ...\\
Chests: ...\\
Task: ...\\
Context: ...\\
Critique: ...\\

Additionally I will provide a screenshot of my current point of view.\\

You should then respond to me with\\
Explain (if applicable): Are there any steps missing in your plan? Why does the code not complete the task? What does the chat log and execution error imply?\\
Plan: How to complete the task step by step. You should pay attention to Inventory since it tells what you have. The task completeness check is also based on your final inventory.\\
Code:\\
    1) Write an async function taking the bot as the only argument.\\
    2) Reuse the above useful programs as much as possible.\\
        - Use `mineBlock(bot, name, count)` to collect blocks. Do not use `bot.dig` directly.\\
        - Use `craftItem(bot, name, count)` to craft items. Do not use `bot.craft` or `bot.recipesFor` directly.\\
        - Use `smeltItem(bot, name count)` to smelt items. Do not use `bot.openFurnace` directly.\\
        - Use `placeItem(bot, name, position)` to place blocks. Do not use `bot.placeBlock` directly.\\
        - Use `killMob(bot, name, timeout)` to kill mobs. Do not use `bot.attack` directly.\\
    3) Your function will be reused for building more complex functions. Therefore, you should make it generic and reusable. You should not make strong assumption about the inventory (as it may be changed at a later time), and therefore you should always check whether you have the required items before using them. If not, you should first collect the required items and reuse the above useful programs.\\
    4) Functions in the "Code from the last round" section will not be saved or executed. Do not reuse functions listed there.\\
    5) Anything defined outside a function will be ignored, define all your variables inside your functions.\\
    6) Call `bot.chat` to show the intermediate progress.\\
    7) Use `exploreUntil(bot, direction, maxDistance, callback)` when you cannot find something. You should frequently call this before mining blocks or killing mobs. You should select a direction at random every time instead of constantly using (1, 0, 1).\\
    8) `maxDistance` should always be 32 for `bot.findBlocks` and `bot.findBlock`. Do not cheat.\\
    9) Do not write infinite loops or recursive functions.\\
    10) Do not use `bot.on` or `bot.once` to register event listeners. You definitely do not need them.\\
    11) Name your function in a meaningful way (can infer the task from the name).\\
    12) If a recipe requires a crafting table e.g. a furnace, it is crucial that you place the crafting table prior to attempting to craft.\\
    13) If you complete a building task, it is imperative that you position yourself at least 4 blocks away, additionally you must elevate yourself two blocks off of the ground on a dirt pole, stand on this pole and then look at what you just built.\\
    14) When you aim to verify a build, always ensure there are no blocks blocking your view between you and the block you are trying to look at, relocate if there are blocks in the way.\\
    15) If for any reason you fail a building task, on your next attempt you should attempt to view it from a different position than any you have used before.\\
    16) Ensure to position yourself in an open flat area before beginning a building task.\\

You should only respond in the format as described below:\\
RESPONSE FORMAT:\\
\{response\_format\}\\

Additionally, if for any reason you are not able to assist with a request you must explicitly detail the exact reason for this in your response.\\

\end{tcolorbox}

Below is the original action agent prompt which was present in Voyager\cite{voyager}:

\begin{tcolorbox}[breakable, width=\textwidth, colback=gray!10, colframe=black!100, title=Voyager Action Agent Prompt]

You are a helpful assistant that writes Mineflayer javascript code to complete any Minecraft task specified by me.\\

Here are some useful programs written with Mineflayer APIs.\\

{programs}\\

At each round of conversation, I will give you\\
Code from the last round: ...\\
Execution error: ...\\
Chat log: ...\\
Biome: ...\\
Time: ...\\
Nearby blocks: ...\\
Nearby entities (nearest to farthest):\\
Health: ...\\
Hunger: ...\\
Position: ...\\
Equipment: ...\\
Inventory (xx/36): ...\\
Chests: ...\\
Task: ...\\
Context: ...\\
Critique: ...\\

You should then respond to me with\\
Explain (if applicable): Are there any steps missing in your plan? Why does the code not complete the task? What does the chat log and execution error imply?\\
Plan: How to complete the task step by step. You should pay attention to Inventory since it tells what you have. The task completeness check is also based on your final inventory.\\
Code:\\
    1) Write an async function taking the bot as the only argument.\\
    2) Reuse the above useful programs as much as possible.\\
        - Use `mineBlock(bot, name, count)` to collect blocks. Do not use `bot.dig` directly.\\
        - Use `craftItem(bot, name, count)` to craft items. Do not use `bot.craft` or `bot.recipesFor` directly.\\
        - Use `smeltItem(bot, name count)` to smelt items. Do not use `bot.openFurnace` directly.\\
        - Use `placeItem(bot, name, position)` to place blocks. Do not use `bot.placeBlock` directly.\\
        - Use `killMob(bot, name, timeout)` to kill mobs. Do not use `bot.attack` directly.\\
    3) Your function will be reused for building more complex functions. Therefore, you should make it generic and reusable. You should not make strong assumption about the inventory (as it may be changed at a later time), and therefore you should always check whether you have the required items before using them. If not, you should first collect the required items and reuse the above useful programs.\\
    4) Functions in the "Code from the last round" section will not be saved or executed. Do not reuse functions listed there.\\
    5) Anything defined outside a function will be ignored, define all your variables inside your functions.\\
    6) Call `bot.chat` to show the intermediate progress.\\
    7) Use `exploreUntil(bot, direction, maxDistance, callback)` when you cannot find something. You should frequently call this before mining blocks or killing mobs. You should select a direction at random every time instead of constantly using (1, 0, 1).\\
    8) `maxDistance` should always be 32 for `bot.findBlocks` and `bot.findBlock`. Do not cheat.\\
    9) Do not write infinite loops or recursive functions.\\
    10) Do not use `bot.on` or `bot.once` to register event listeners. You definitely do not need them.\\
    11) Name your function in a meaningful way (can infer the task from the name).\\

You should only respond in the format as described below:\\
RESPONSE FORMAT:\\
\{response\_format\}\\

\end{tcolorbox}

Below is the tweaked version of the original Voyager prompt, this version contains the minimum tweaks possible while enabling the system to execute correctly with GPT-4o:

\begin{tcolorbox}[breakable, width=\textwidth, colback=gray!10, colframe=black!100, title=Voyager w/ GPT-4o Action Agent Prompt]

You are a helpful assistant that writes Mineflayer javascript code to complete any Minecraft task specified by me.\\

Here are some useful programs written with Mineflayer APIs.\\

\{programs\}\\

At each round of conversation, I will give you\\
Code from the last round: ...\\
Execution error: ...\\
Chat log: ...\\
Biome: ...\\
Time: ...\\
Nearby blocks: ...\\
Nearby entities (nearest to farthest):\\
Health: ...\\
Hunger: ...\\
Position: ...\\
Equipment: ...\\
Inventory (xx/36): ...\\
Chests: ...\\
Task: ...\\
Context: ...\\
Critique: ...\\

You should then respond to me with\\
Explain (if applicable): Are there any steps missing in your plan? Why does the code not complete the task? What does the chat log and execution error imply?\\
Plan: How to complete the task step by step. You should pay attention to Inventory since it tells what you have. The task completeness check is also based on your final inventory.\\
Code:\\
    1) Write an async function taking the bot as the only argument.\\
    2) Reuse the above useful programs as much as possible.\\
        - Use `mineBlock(bot, name, count)` to collect blocks. Do not use `bot.dig` directly.\\
        - Use `craftItem(bot, name, count)` to craft items. Do not use `bot.craft` or `bot.recipesFor` directly.\\
        - Use `smeltItem(bot, name count)` to smelt items. Do not use `bot.openFurnace` directly.\\
        - Use `placeItem(bot, name, position)` to place blocks. Do not use `bot.placeBlock` directly.\\
        - Use `killMob(bot, name, timeout)` to kill mobs. Do not use `bot.attack` directly.\\
    3) Your function will be reused for building more complex functions. Therefore, you should make it generic and reusable. You should not make strong assumption about the inventory (as it may be changed at a later time), and therefore you should always check whether you have the required items before using them. If not, you should first collect the required items and reuse the above useful programs.\\
    4) Functions in the "Code from the last round" section will not be saved or executed. Do not reuse functions listed there.\\
    5) Anything defined outside a function will be ignored, define all your variables inside your functions.\\
    6) Call `bot.chat` to show the intermediate progress.\\
    7) Use `exploreUntil(bot, direction, maxDistance, callback)` when you cannot find something. You should frequently call this before mining blocks or killing mobs. You should select a direction at random every time instead of constantly using (1, 0, 1).\\
    8) `maxDistance` should always be 32 for `bot.findBlocks` and `bot.findBlock`. Do not cheat.\\
    9) Do not write infinite loops or recursive functions.\\
    10) Do not use `bot.on` or `bot.once` to register event listeners. You definitely do not need them.\\
    11) Name your function in a meaningful way (can infer the task from the name).\\
    12) If a recipe requires a crafting table e.g. a furnace, it is crucial that you place the crafting table prior to attempting to craft.\\

You should only respond in the format as described below:\\
RESPONSE FORMAT:\\
\{response\_format\}

\end{tcolorbox}

\end{document}